\documentclass[11pt]{article}
\usepackage{acl}
\usepackage{times}
\usepackage{latexsym}
\usepackage[T1]{fontenc}
\usepackage[utf8]{inputenc}
\usepackage{microtype}
\usepackage{amsmath}
\usepackage{amssymb}
\usepackage{graphicx}
\usepackage{booktabs}
\usepackage{multirow}
\usepackage{array}
\usepackage[table]{xcolor}
\usepackage{colortbl}
\usepackage{makecell}
\usepackage{tabularx}
\usepackage{pifont}
\usepackage{tikz}
\usepackage{pgfplots}
\pgfplotsset{compat=1.18}
\usetikzlibrary{shapes.geometric, arrows.meta, positioning, fit, calc, backgrounds}
\usepackage{hyperref}
\usepackage{fontawesome5}

\definecolor{benchV}{HTML}{F5C6A0}   
\definecolor{benchB}{HTML}{C8B6E2}   
\definecolor{benchN}{HTML}{A8D5BA}   
\definecolor{benchD}{HTML}{F4B6B6}   
\definecolor{benchC}{HTML}{9FD8E8}   
\definecolor{benchX}{HTML}{D4D4D4}   
\definecolor{benchO}{HTML}{F5D76E}   
\definecolor{benchR}{HTML}{E8B8D0}   
\definecolor{benchI}{HTML}{B8D4E8}   
\definecolor{benchW}{HTML}{F2C0DC}   
\definecolor{benchP}{HTML}{C4D8B0}   
\definecolor{benchA}{HTML}{E0CDA9}   
\definecolor{benchG}{HTML}{D0B8A8}   

\definecolor{effLat}{HTML}{FFEBF9}    
\definecolor{effParam}{HTML}{E9FFEF}  
\definecolor{effMem}{HTML}{F1E1FF}    
\definecolor{effFlops}{HTML}{FFECE1}  
\definecolor{effToken}{HTML}{E9FFFE}  

\definecolor{secthead}{HTML}{FFFEEC}  
\definecolor{stripebg}{HTML}{FAFAFA}  
\definecolor{citecol}{HTML}{1F7F9C}   
\definecolor{rulecol}{HTML}{707070}   

\definecolor{Lone}{HTML}{CAFFBF}   
\definecolor{Ltwo}{HTML}{CAFFBF}   
\definecolor{Lthree}{HTML}{CAFFBF} 
\definecolor{Lfour}{HTML}{CAFFBF}  
\providecommand{\bench}[2]{{\colorbox{#1}{\textbf{\scriptsize\sffamily #2}}}}
\providecommand{\bV}{\bench{benchV}{V}}
\providecommand{\bB}{\bench{benchB}{B}}
\providecommand{\bN}{\bench{benchN}{N}}
\providecommand{\bD}{\bench{benchD}{D}}
\providecommand{\bC}{\bench{benchC}{C}}

\providecommand{\bR}{\bench{benchR}{R}}
\providecommand{\bI}{\bench{benchI}{I}}
\providecommand{\bW}{\bench{benchW}{W}}
\providecommand{\bP}{\bench{benchP}{P}}
\providecommand{\bA}{\bench{benchA}{A}}

\providecommand{\efftag}[2]{{\colorbox{#1}{\scriptsize\sffamily #2}}}
\providecommand{\eLat}{\efftag{effLat}{Latency\,$\downarrow$}}
\providecommand{\eParam}{\efftag{effParam}{Parameters\,$\downarrow$}}
\providecommand{\eMem}{\efftag{effMem}{Memory\,$\downarrow$}}
\providecommand{\eFlops}{\efftag{effFlops}{FLOPs\,$\downarrow$}}

\providecommand{\eToken}{\efftag{effToken}{Tokens\,$\downarrow$}}
\providecommand{\Ltag}[2]{%
  {\colorbox{#1}{\scriptsize\sffamily\bfseries #2}}%
}
\providecommand{\Lo}{\Ltag{Lone}{L1}}
\providecommand{\Lt}{\Ltag{Ltwo}{L2}}
\providecommand{\Lth}{\Ltag{Lthree}{L3}}
\providecommand{\Lf}{\Ltag{Lfour}{L4}}

\providecommand{\cciteone}[1]{{\color{citecol}\cite{#1}}}

\providecommand{\opencode}[1]{\href{#1}{\faGlobe}}

\newcolumntype{Y}{>{\raggedright\arraybackslash}X}

\providecommand{\secthrow}[2]{%
  \rowcolor{secthead}\multicolumn{5}{@{}l@{}}{%
    \rule{0pt}{2.8ex}%
    \quad\textbullet\;\textbf{~#1}~\textit{#2}%
    \rule[-1.2ex]{0pt}{0pt}%
  }\\
}

\newcommand{\cnum}[2]{%
\tikz[baseline=(char.base)]{
\node[shape=circle, fill=#1, text=white, inner sep=1pt] (char) {#2};
}}

\definecolor{morandiBlue}{RGB}{130,150,170}
\definecolor{morandiPurple}{RGB}{160,140,170}
\definecolor{morandiRed}{RGB}{180,120,120}
\definecolor{myBlue}{HTML}{A2D2FF}
\definecolor{myPink}{HTML}{FFAFCC}
\definecolor{myPurple}{HTML}{E0AAFF}
\definecolor{myGreen}{HTML}{70E000}
\newcommand{\autocnum}[1]{%
\ifcase#1
\or \cnum{myBlue}{#1}
\or \cnum{myPink}{#1}
\or \cnum{myPurple}{#1}
\or \cnum{myGreen}{#1}
\else \cnum{myBlue}{#1}
\fi
}
\newcolumntype{s}{>{\hsize=.62\hsize\raggedright\arraybackslash}X}
\newcolumntype{d}{>{\hsize=1.38\hsize\raggedright\arraybackslash}X}

\usepackage{tgstyle}

\usepackage{calligra}

\definecolor{deepblue}{HTML}{2B4F81}

\title{
  \raisebox{-0.2\height}{\includegraphics[height=2em]{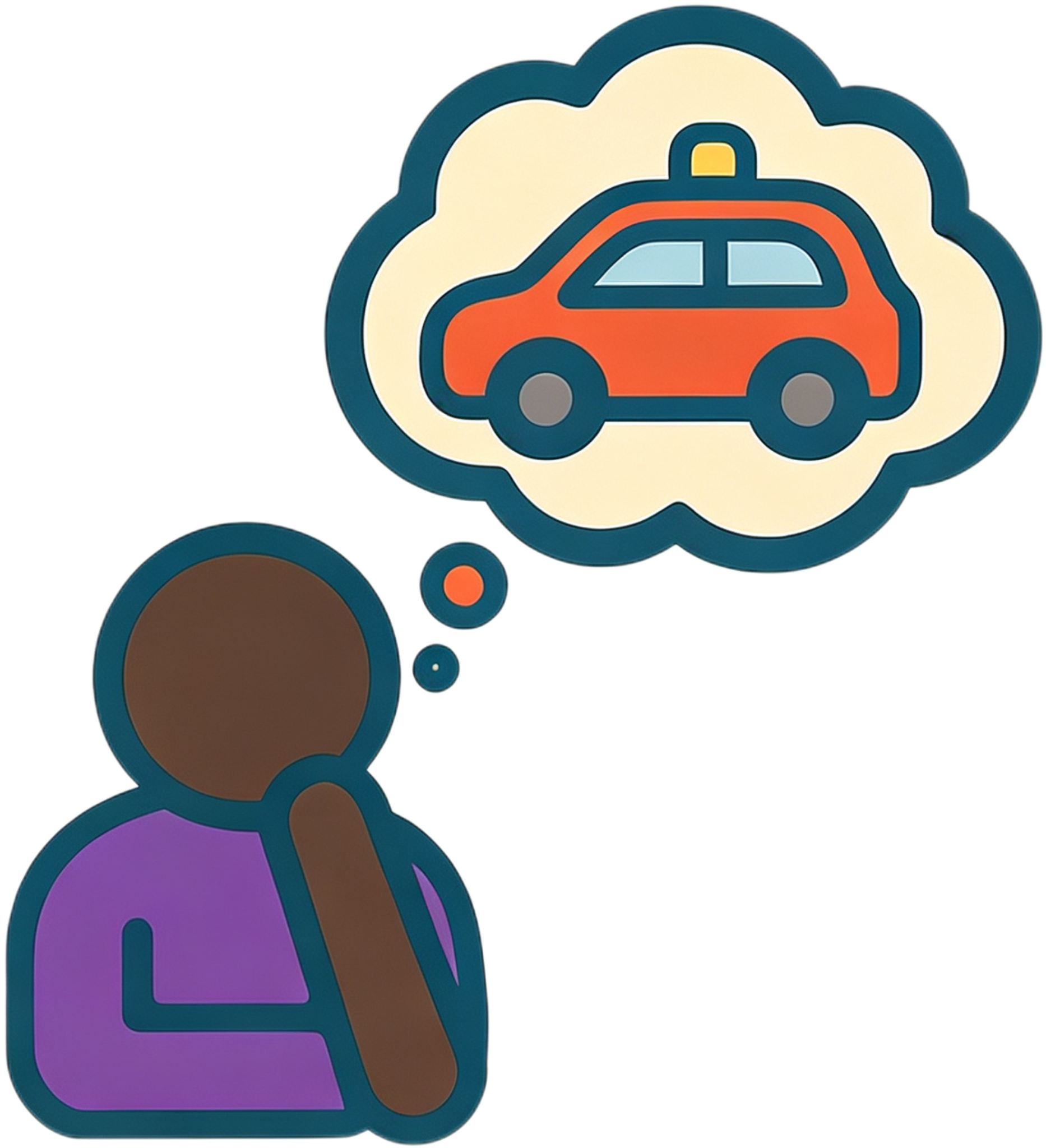}}%
  \hspace{0.5em}
Rethinking Language’s Role in Efficient VLA for Autonomous Vehicles: Toward Smarter, Trustworthy Driving}

\author{Tongfei Guo$^{1}$ and Lili Su$^{1}$%
\thanks{This work was supported in part by the NSF CAREER award under grant~2340482 and by NSF award~2414652. $^{1}$Department of Electrical and Computer Engineering, Northeastern University, Boston, MA, USA.
        {\tt\small \{guo.t, l.su\}@northeastern.edu}
        }%
}

\begin{document}
\maketitle

\begin{abstract}
  Vision--Language--Action (VLA) models are reshaping autonomous driving (AD) by
  unifying perception, reasoning, and control through language, enabling
  semantic grounding, interpretable decisions, and better long-tail generalization. But language is expensive onboard: latency and memory budgets are tight, and autoregressive decoding is inherently sequential.
 We reframe the central question as \emph{when and where} language should act at inference, since inference cost recurs at every deployed frame while training cost is paid once.
  We introduce the \textit{Language Residue taxonomy} to organize
  methods by their inference-time use of language: train-time-only supervision~(\Lo), latent non-textual
  reasoning~(\Lt), conditional invocation~(\Lth), and full per-frame
  generation~(\Lf).
  We review representative methods and tag each
  across five deployment axes (\eLat, \eParam, \eMem, \eFlops, \eToken), analyzing them on major open- and closed-loop driving benchmarks (e.g., nuScenes, NAVSIM, Bench2Drive).
  We further trace how efficient methods from NLP/LLM are adapted in AD, identifying the constraints and motivations driving these adaptations. A continuously updated repository will be available at 
  \href{https://github.com/iamtongfei/Awesome-Efficient-VLA4AD}{GitHub}.
  \end{abstract}

\section{Introduction}
\label{sec:intro}
Vision-Language-Action (VLA) models have emerged as a compelling paradigm for
autonomous driving~(AD), extending end-to-end autonomous driving 
with contextual reasoning 
\citep{drivelm2023,drivegpt42023,drivevlm2024,senna2024,lmdrive2024}.
The promise is clear: language can 
mitigate the uncertainties in long-tail scenarios, and align driving behavior with passenger preferences (e.g.~safety vs.~comfort vs.~speeds). 
Deploying language under hard deadlines exposes a trade-off between its semantic
leverage and its inference cost: AD control loops must close within roughly
$20$--$100$\,ms ($10$--$50$\,Hz), yet one autoregressive VLM inference on typical
onboard hardware costs $500$--$2000$\,ms~\citep{drivevlmrl2026}.

\begin{figure}[htbp]
    \centering
    \includegraphics[width=1\linewidth]{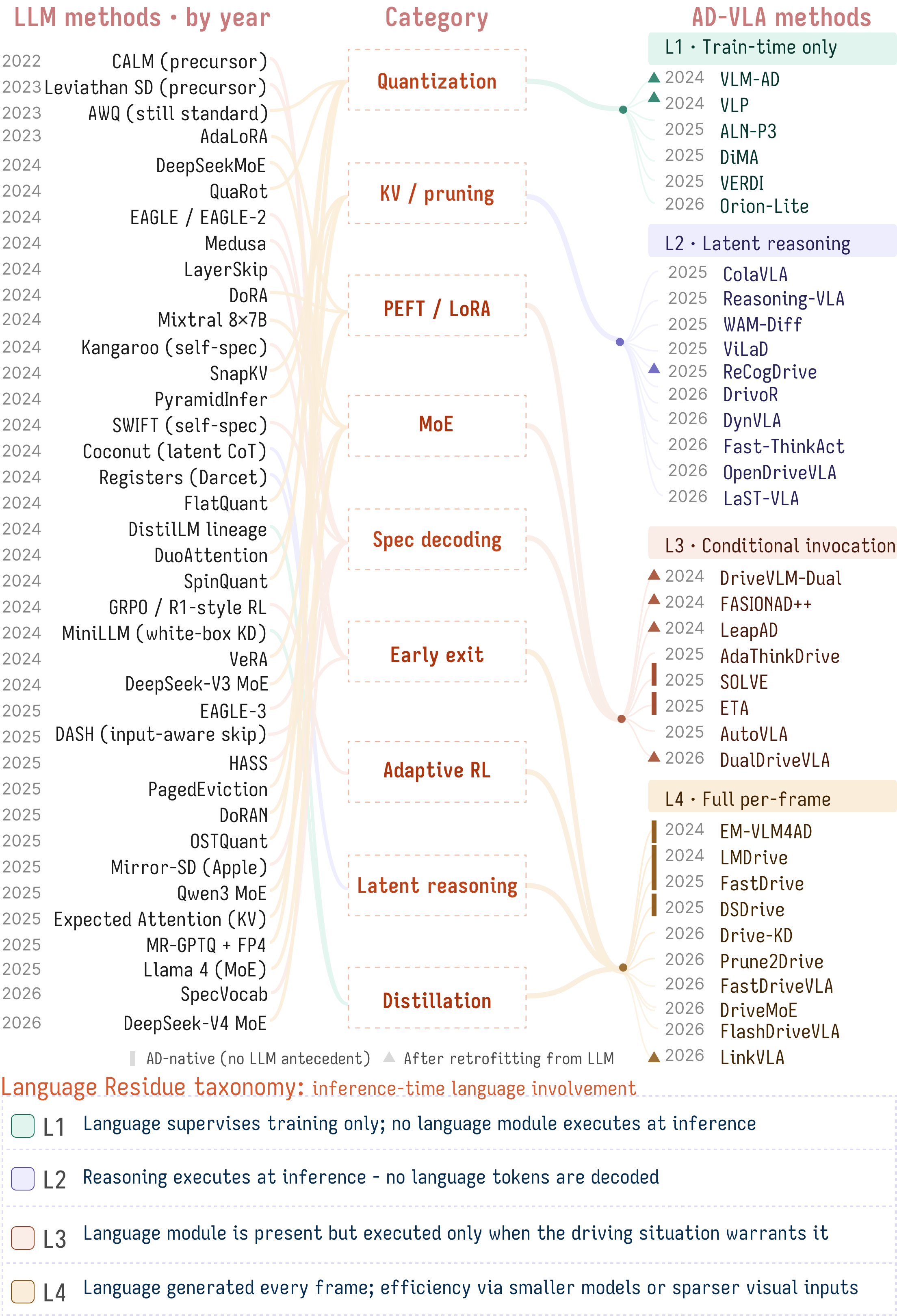}
    \caption{%
    NLP/LLM efficiency models ({\it left}) routed through technique categories ({\it middle}) into their AD-VLA adaptation ({\it right}), grouped by Language Residue level L1–L4.}
    \label{fig:migration}
    \vspace{-1em}
\end{figure}

%
%
%

\providecolor{statusYes}{HTML}{2E8B3D}     
\providecolor{statusNo}{HTML}{C0392B}      
\providecolor{statusPartial}{HTML}{E67E22} 

\providecommand{\yes}{\textcolor{statusYes}{\ding{51}}}
\providecommand{\no}{\textcolor{statusNo}{\ding{55}}}
\providecommand{\yescell}{\yes}
\providecommand{\nocell}{\no}
\providecommand{\partialcell}{\textcolor{statusPartial}{\ensuremath{\bigtriangleup}}}
\providecommand{\nascope}{--}

\providecolor{techDst}{HTML}{1F77B4}   
\providecolor{techPr} {HTML}{8E44AD}   
\providecolor{techDS} {HTML}{0E7C7B}   
\providecolor{techQt} {HTML}{8C564B}   
\providecolor{techMoE}{HTML}{C71585}   
\providecolor{techSpA}{HTML}{2C3E50}   
\providecolor{techKV} {HTML}{B8860B}   
\providecolor{techLoR}{HTML}{4B0082}   
\providecolor{techEE} {HTML}{006D5B}   
\providecolor{techSpD}{HTML}{555555}   

\providecommand{\tkDst}{\textcolor{techDst}{\textbf{Dst}}}
\providecommand{\tkPr} {\textcolor{techPr} {\textbf{Pr}}}
\providecommand{\tkDS} {\textcolor{techDS} {\textbf{DS}}}
\providecommand{\tkQt} {\textcolor{techQt} {\textbf{Qt}}}
\providecommand{\tkMoE}{\textcolor{techMoE}{\textbf{MoE}}}
\providecommand{\tkSpA}{\textcolor{techSpA}{\textbf{SpA}}}
\providecommand{\tkKV} {\textcolor{techKV} {\textbf{KV}}}
\providecommand{\tkLoR}{\textcolor{techLoR}{\textbf{LoR}}}
\providecommand{\tkEE} {\textcolor{techEE} {\textbf{EE}}}
\providecommand{\tkSpD}{\textcolor{techSpD}{\textbf{SpD}}}

\begin{table*}[tbhp]
\centering
\scriptsize
\setlength{\tabcolsep}{4pt}
\renewcommand{\arraystretch}{1.15}
\caption{%
  \textbf{Comparison of existing surveys on efficient-method technique coverage.}
  Eff. = efficiency as the main focus.
  Tech-keys:
  \tkDst{}=knowledge distillation,
  \tkPr{}=pruning / token reduction,
  \tkDS{}=dual-system (fast/slow) inference,
  \tkQt{}=quantization,
  \tkMoE{}=mixture-of-experts,
  \tkSpA{}=sparse / linear attention,
  \tkKV{}=KV-cache optimization,
  \tkLoR{}=LoRA / PEFT,
  \tkEE{}=early-exit,
  \tkSpD{}=speculative decoding.
  Status: \yes{}~= primary coverage, \partialcell{}~= partial / mentioned, \no{}~= absent, CLE = Closed loop evaluation.
}
\label{tab:survey_comparison}
\begin{tabular}{@{}l l c cccccccccc c@{}}
\toprule
\textbf{Survey} & \textbf{Organizing axis}
& \makecell{\textbf{Eff.}}
& \tkDst & \tkPr & \tkDS & \tkQt & \tkMoE
& \tkSpA & \tkKV & \tkLoR & \tkEE & \tkSpD
& \makecell{\textbf{CLE}} \\
\midrule


\citet{yang2023survey}          & Task taxonomy (QA / planning / generation)
& \nocell
& \nocell & \nocell & \nocell & \nocell & \nocell
& \nocell & \nocell & \nocell & \nocell & \nocell
& \partialcell \\

\citet{cui2024llm4ad} & Application domain (perception / planning / HMI)
& \partialcell
& \nocell & \nocell & \partialcell & \partialcell & \nocell
& \nocell & \nocell & \partialcell & \nocell & \nocell
& \partialcell \\

\citet{zhou2024tiv}     & VFM role in AD stack
& \partialcell
& \partialcell & \partialcell & \nocell & \partialcell & \nocell
& \nocell & \nocell & \partialcell & \nocell & \nocell
& \partialcell \\

\citet{chen2024end}            & Pipeline placement (modular vs.\ E2E)
& \nocell
& \nocell & \nocell & \nocell & \nocell & \nocell
& \nocell & \nocell & \nocell & \nocell & \nocell
& \yescell \\

\citet{xu2024ondevice}          & On-device LM deployment
& \partialcell
& \yescell & \yescell & \partialcell & \yescell & \yescell
& \partialcell & \partialcell & \partialcell & \nocell & \partialcell
& \nocell \\

\citet{wang2024slmsurvey} & Small language models taxonomy
& \partialcell
& \yescell & \yescell & \nocell & \yescell & \partialcell
& \partialcell & \partialcell & \yescell & \partialcell & \nocell
& \nocell \\

\citet{hu2026vlaad}             & Runtime arch. (E2E vs.\ dual-system)
& \partialcell
& \nocell & \nocell & \yescell & \nocell & \partialcell
& \partialcell & \nocell & \nocell & \nocell & \nocell
& \yescell \\

\citet{yu2025effvla} & Efficiency pipeline (model / training / data)
& \yescell
& \yescell & \yescell & \yescell & \yescell & \yescell
& \yescell & \partialcell & \yescell & \partialcell & \yescell
& \nocell \\

\citet{sapkota2025}             & VLA design (architecture + training)
& \partialcell
& \nocell & \partialcell & \partialcell & \partialcell & \partialcell
& \nocell & \partialcell & \partialcell & \nocell & \nocell
& \nocell \\

\citet{sharshar2025edgevlm}     & Edge deployment (VLM systems)
& \yescell
& \yescell & \yescell & \nocell & \yescell & \partialcell
& \nocell & \nocell & \partialcell & \nocell & \nocell
& \nocell \\

\citet{jiang2025vla4ad} & VLA building blocks + evolution
& \partialcell
& \partialcell & \partialcell & \partialcell & \partialcell & \partialcell
& \nocell & \nocell & \partialcell & \nocell & \nocell
& \yescell \\

\citet{zhu2025survey}           & Pipeline dichotomy (modular vs.\ E2E)
& \partialcell
& \nocell & \nocell & \nocell & \nocell & \nocell
& \nocell & \nocell & \nocell & \nocell & \nocell
& \partialcell \\

\citet{cai2024moesurvey} & Architecture class (Mixture-of-Experts)
& \partialcell
& \nocell & \nocell & \nocell & \nocell & \yescell
& \nocell & \nocell & \partialcell & \nocell & \nocell
& \nocell \\
\midrule
\rowcolor{yellow!18}
\textbf{Ours} & \textbf{Inference-lifecycle language role}
& \yescell
& \yescell & \yescell & \yescell & \yescell & \yescell
& \partialcell & \yescell & \yescell & \yescell & \yescell
& \yescell \\

\bottomrule
\end{tabular}
\end{table*}

Recent advances in efficient VLA4AD\footnote{Vision--Language--Action models for Autonomous Driving, introduced by \citet{jiang2025vla4ad}.}
demonstrate that scale alone does not determine capability: careful language allocation at smaller scales can match or surpass substantially larger baselines.
Orion-Lite \citep{orionlite2026} distills a 7B teacher into a 0.1B vision-only
student that surpasses its teacher on Bench2Drive~\citep{jia2024bench2drive}
while bypassing the language backbone at inference. DualDriveVLA
\citep{dualdrivevla2026} invokes its language module on only about 15\% of
scenarios and still matches always-on VLM accuracy on NAVSIM v1~\citep{dauner2024navsim}.
Despite unrelated architectures, both \emph{reduce} the language active at inference
while preserving its function: reasoning, grounding, and long-tail generalization.

\begin{figure*}[t]
    \centering
    \includegraphics[width=0.94\linewidth]{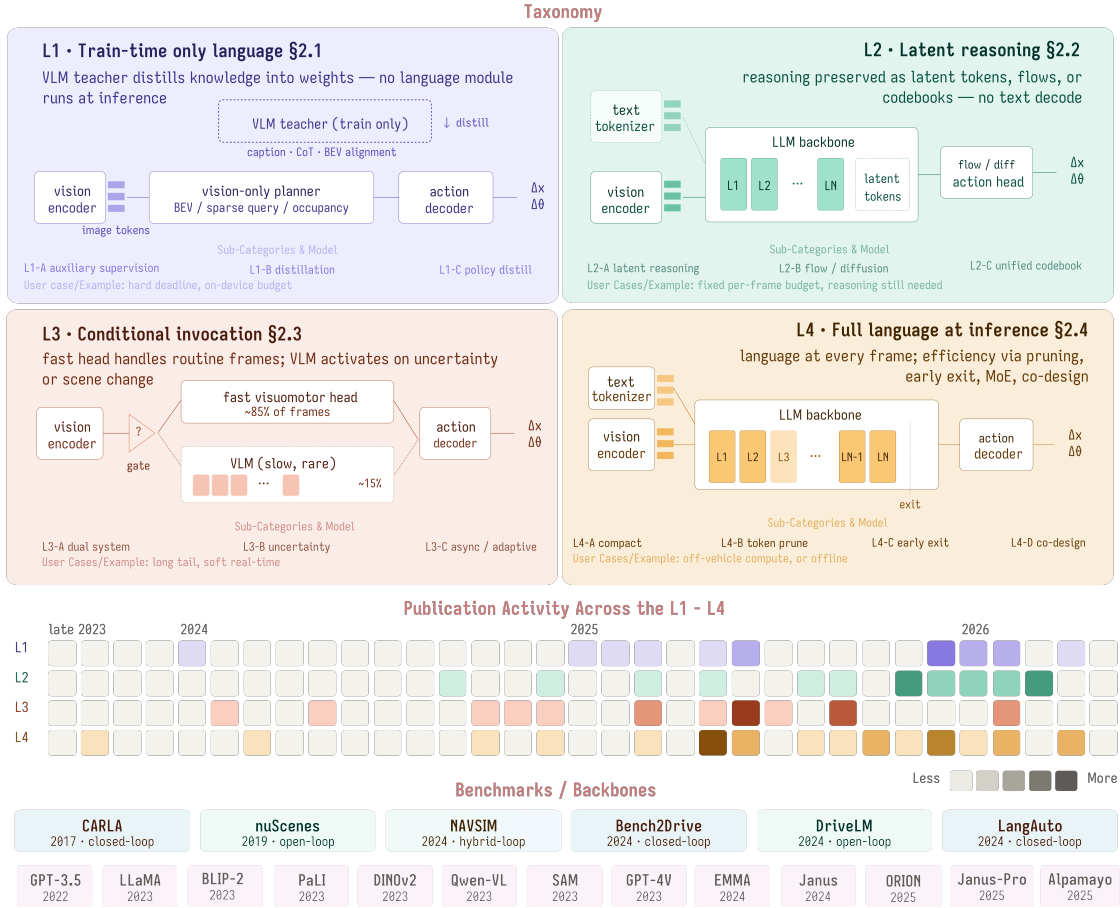}
    \caption{{A taxonomy of efficient VLA4AD pipeline.
    Sensor inputs (multi-camera images, lidar) feed a vision encoder; outputs are consumed
    by a language backbone that issues planning decisions; an action decoder converts plans
    into vehicle controls.
    The Language Residue taxonomy classifies how much language is active at inference,
    from \Lf\ (full per-frame autoregressive generation) down to \Lo\ (no language at runtime).
    Deployment cost concentrates at the language backbone, motivating the allocation question
    addressed here.
    }}
    \label{fig:vla-overview}
\end{figure*}

We therefore reframe the design question: not \emph{how small} a model should be, but \emph{when and where language should participate at inference}, since training occurs only once while inference cost is paid at every deployed frame.
To organize methods along this axis, we introduce the \textit{Language Residue taxonomy}, which categorizes representative methods by their inference-time use of language: train-time-only supervision~(\Lo), latent non-textual reasoning~(\Lt), conditional invocation~(\Lth), and full per-frame generation~(\Lf). Figure~\ref{fig:vla-overview} presents the full taxonomy.

The taxonomy spans the major contemporary approaches, which differ in what
becomes of language at inference:
{\it \ding{172} Knowledge distillation}~\citep{dima2025,orionlite2026}
relocates its value into the student's weights.
{\it \ding{173} Language-supervised pretraining}~\citep{vlp2024,vlmad2025,alnp32025}
relocates it into learned feature geometry.
{\it \ding{174} Latent world models and flow-matching planners}~\citep{colavla2025,lastvla2026,wamflow2025,goalflow2025}
encode reasoning in continuous representations, dropping the token vocabulary
altogether.
{\it \ding{175} Conditional dispatch}~\citep{drivevlm2024,dualdrivevla2026,eta2025}
invokes language only when a frame demands it: scene-gated routers and
asynchronous dual-systems send routine frames down a lightweight reactive path.
{\it \ding{176} Within-model efficiency} keeps language active every frame but
cuts its cost: mixture-of-experts (MoE) architectures activate only a subset of
experts per token~\citep{drivemoe2025}, while quantization, token pruning, and
co-designed serving stacks~\citep{stprune2026,flashdrivevla2026} trim latency and
memory without altering language's temporal role.

Prior surveys (Table~\ref{tab:survey_comparison}) either catalog AD--LLM
capabilities~\citep{yang2023survey,cui2024llm4ad,zhou2024tiv} or
organize VLA methods by
architecture~\citep{jiang2025vla4ad,sapkota2025,hu2026vlaad}. The
closest, \citet{yu2025effvla} treats efficiency as model compression
borrowed from the general LLM~\citep{wang2024slmsurvey,xu2024ondevice}
and edge-VLM~\citep{sharshar2025edgevlm} literature. 
It leaves unexamined how much language a deployed system must run at inference, and what determines that allocation in the first place.

\begin{figure}
    \centering
    \includegraphics[width=1\linewidth]{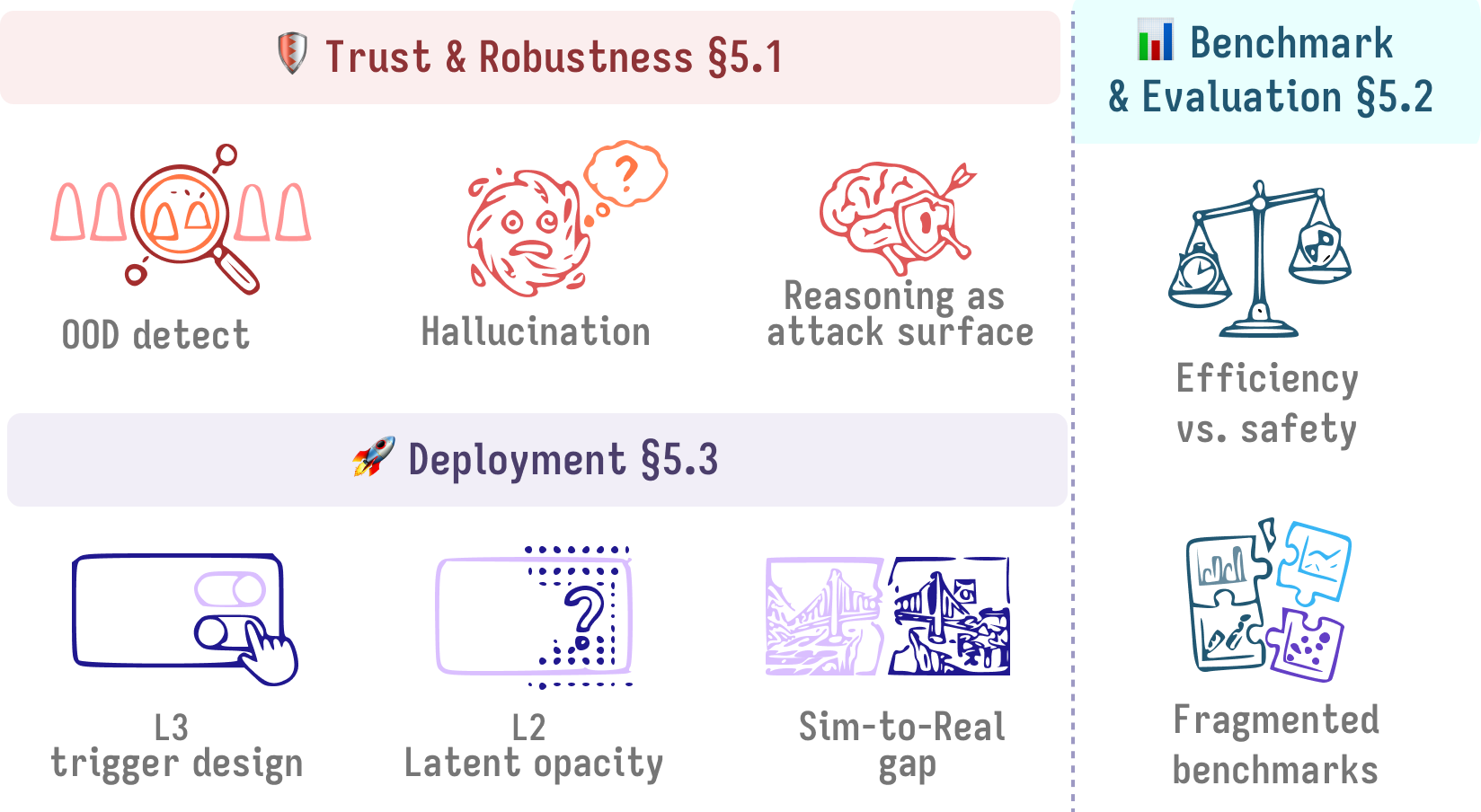}
    \caption{{Open problems in efficient VLA4AD.} Three tiers of gaps: trust and
robustness, benchmark and evaluation, and deployment.}
\label{fig:open-problems}
\vspace{-1em}
\end{figure}

We fill that gap by rethinking language as an
allocation decision: \emph{retain what language contributes,
eliminate what it costs}.
We review recent methods (Section~\ref{sec:taxonomy}), annotating each across five deployment axes
(\eLat, \eParam, \eMem, \eFlops, \eToken), and analyze them on major
open- and closed-loop driving benchmarks (Section~\ref{sec:benchmarks}). 
We further trace how efficient methods from NLP/LLM systems are adapted in AD
(Section~\ref{sec:nlp-ad}), identifying the constraints and motivations that
reshape each under hard-deadline\footnote{A fixed control-loop cycle time that
cannot be exceeded.} 
deployment.
We close with the open problems that remain
(Section~\ref{sec:saturation}, Figure~\ref{fig:open-problems}).
We hope this survey, together with our continuously updated repository at
\href{https://github.com/iamtongfei/Awesome-Efficient-VLA4AD}{\texttt{GitHub}},
helps researchers and practitioners navigate the rapidly evolving landscape and
inspires further advances in efficient VLA design.

\section{Language's Role in VLA}
\label{sec:taxonomy}

\begin{table*}[t]
\centering
\scriptsize
\setlength{\tabcolsep}{4pt}
\renewcommand{\arraystretch}{1.3}
\arrayrulecolor{rulecol}

\caption{%
  \textbf{L1 --- Train-Time-Only Language Methods.}
  Language supervises training; no language module executes at inference.
  \textbullet\ \textbf{Benchmarks:}
  \bB~Bench2Drive~\cciteone{jia2024bench2drive},
  \bN~nuScenes~\cciteone{caesar2020nuscenes},
  \bD~DriveLM~\cciteone{drivelm2023}.
  \textbullet\ \faGlobe\ links to an official open-source repository or project page.
}
\label{tab:l1_style}

\rowcolors{2}{white}{stripebg}
\begin{tabularx}{\textwidth}{@{} >{\centering\arraybackslash}p{0.4cm} Y Y Y >{\raggedright\arraybackslash}p{1.6cm} @{}}
\toprule
\textbf{\#} & \textbf{Method} & \textbf{Backbone / Params} & \textbf{Efficiency} & \textbf{Bench.} \\
\midrule

\secthrow{A}{Auxiliary Supervision --- language shapes gradients; inference is purely visuomotor}
1  & VLP~\cite{vlp2024}             & UniAD/VAD + text aux.        & \eLat\,\eParam   & \bN     \\
2  & VLM-AD~\cite{vlmad2025}        & UniAD + GPT-4o                & \eLat\,\eParam   & \bN     \\
3  & VLM-E2E~\cite{liu2025vlme2e}   & E2E + driver attention       & \eLat            & \bN     \\
4  & OccVLA~\cite{occvla2025}       & VLM + implicit 3D occupancy    & \eFlops          & \bN     \\

\secthrow{B}{Feature Distillation --- VLM teacher transfers representations to a compact student}
5  & RSD~\cite{qin2025rsd}          & E2E + VLM risk, $\sim$50M    & \eParam          & \bB     \\
6  & ALN-P3~\cite{alnp32025}        & BEV + LLM-aligned P3         & \eLat            & \bN     \\
7  & VERDI~\cite{feng2025verdi}     & VLM-embedded reasoning       & \eLat   & \bN     \\
8  & DiMA~\cite{dima2025}                       & VAD + MLLM $\to$ VAD-Tiny/Base student & \eLat   & \bN~\bD \\
9  & Orion-Lite \opencode{https://github.com/tue-mps/Orion-Lite}~\cite{orionlite2026}            & ORION 7B $\to$ vision student  & \eLat\,\eMem  & \bB     \\



\bottomrule
\end{tabularx}

\vspace{2pt}
{\tiny
\textit{L1 distinguishing property:} deploying these models incurs \emph{zero} language-module cost at inference,
efficiency gains are fully amortized into weights during training.
}
\end{table*}

\subsection{Train-Time-Only Language (L1)}
\label{sec:l1} L1 methods remove language from the inference graph entirely: language
participates only during training, and the deployed model is a purely visuomotor
policy. What varies within L1 is how the language-derived prior reaches the
vision-only weights (Table~\ref{tab:l1_style}).

\emph{Auxiliary supervision} (L1-A) folds language objectives into the
training loss, shaping the gradient without teacher-student distillation. Language thus survives only as a training signal, but that signal is what
transfers language-model reasoning into the vision network, buying zero-shot
generalization to unseen cities and long-tail maneuvers that a vision-only
planner lacks~\citep{vlp2024}.
\emph{Feature distillation} (L1-B) transfers a VLM's knowledge into a compact planner through a shared scene encoder and surrogate tasks. Because the teacher is discarded after training, the driving path runs
language-free yet retains the multimodal LLM's world knowledge in the planner's features,
recovering robustness on rare events such as three-point turns and overtaking at
no inference cost~\citep{dima2025}.
Above mechanisms trade supervision fidelity against deployment simplicity differently, and none yet dominates across benchmarks.

\subsection{Non-Textual Reasoning (L2)}
\label{sec:l2}
L2 methods retain reasoning at inference while replacing its textual surface form with
continuous latent tokens, carrying intermediate cognition without decoding language.
The diversity of L2 designs reflects an unresolved question: which latent substrate best preserves reasoning once explicit text is removed (Table~\ref{tab:l2_styled}).

\emph{Continuous latent reasoning} (L2-A) replaces text decoding with continuous
hidden states to skip generation latency and capture graded uncertainty that
discrete tokens cannot~\citep{bai2026laravla}. 
\emph{Flow- and diffusion-based action heads} (L2-B) replace textual deliberation
with a generative distribution over trajectories, recovering the multimodal choice between overtaking and following that unimodal planners lose~\citep{goalflow2025}. \emph{Unified language--action codebooks} (L2-C) absorb language rather than
discard it, interleaving action and text tokens in a single autoregressive
sequence. Actions are generated directly, without intermediate rationales, while
the LLM's world knowledge and instruction-following remain fully intact.

Across these directions, supervision quality proves more decisive than the
choice of substrate. \emph{LaST-VLA}~\citep{lastvla2026} isolates the principal
failure mode: without explicit supervision, latent tokens degenerate into
trivial representations. Its spatio-temporal supervision aligned with physical
priors reaches PDMS~91.3 on NAVSIM~v1, advancing the state of the art in
planning accuracy, while other methods in the tier report the 5--10$\times$
speedups over textual CoT that make the level attractive as a whole (e.g.,
LaRA-VLA's 90\% latency reduction~\citep{bai2026laravla}). LaST-VLA's own
paper reports no latency or throughput figure.

\begin{table*}[t]
\centering
\scriptsize
\setlength{\tabcolsep}{4pt}
\renewcommand{\arraystretch}{1.3}
\arrayrulecolor{rulecol}

\caption{%
  \textbf{L2 --- Latent Reasoning Methods.}
  Reasoning executes at inference: no language tokens are decoded.
  \textbullet\ \faGlobe\ links to an official open-source repository or project page.
    \textbullet\ \textbf{Benchmarks:} \bV~NAVSIM~\cciteone{dauner2024navsim}, \bB~Bench2Drive~\cciteone{jia2024bench2drive}, \bN~nuScenes~\cciteone{caesar2020nuscenes}, \bI~LIBERO~\cciteone{liu2023libero}/SimplerEnv~\cciteone{li2024simplerenv}, \bR~Robotics, \bA~PhysicalAI-AV~\cciteone{physicalaiav2025}.
}
\label{tab:l2_styled}

\rowcolors{2}{white}{stripebg}
\begin{tabularx}{\textwidth}{@{} >{\centering\arraybackslash}p{0.4cm} Y Y Y >{\raggedright\arraybackslash}p{1.8cm} @{}}
\toprule
\textbf{\#} & \textbf{Method} & \textbf{Backbone / Params} & \textbf{Efficiency} & \textbf{Bench.} \\
\midrule
\secthrow{A}{Latent Reasoning --- continuous or structured latent tokens replace textual chain-of-thought}
1  & LCDrive~\cite{lcdrive2025}                & Qwen3-0.5B + world model         & \eToken\,\eLat         & \bA     \\
2  & Reasoning-VLA~\cite{reasoningvla2025}     & VLA + parallel action queries    & \eLat\,\eToken         & \bN~\bV \\
3  & LaRA-VLA \opencode{https://github.com/LoveJu1y/LaRA-VLA}~\cite{bai2026laravla}            & VLA + latent thinking + pred.    & \eToken\,\eLat         & \bI~\bR \\
4  & LaST-VLA \opencode{https://github.com/luo-yc17/LaST-VLA}~\cite{lastvla2026}               & VLM + ST-latent CoT (3D-FM+WM)   & \eLat\,\eToken         & \bV     \\
5  & DynVLA \opencode{https://yaoyao-jpg.github.io/dynvla}~\cite{dynvla2026}                  & VLA + Dynamics Tokenizer         & \eLat\,\eToken         & \bV~\bB \\
6  & NetRoller \opencode{https://github.com/Rex-sys-hk/NetRoller}~\cite{netroller2025}            & General + specialized interface  & \eLat                  & \bN     \\
7  & HiST-VLA~\cite{histvla2026}               & LLaVA-v1.5-7B (CLIP ViT-L/14 + Vicuna-7B) & \eFlops         & \bV     \\
8  & ColaVLA \opencode{https://github.com/pqh22/ColaVLA}~\cite{colavla2025}                & LLaMA-7B + CLR + HPP             & \eLat\,\eToken         & \bN     \\
9  & DrivoR \opencode{https://github.com/valeoai/DrivoR/}~\cite{drivor2026}                  & ViT + register tokens (Valeo)    & \eLat                  & \bV     \\

\secthrow{B}{Flow / Diffusion Latent Inference --- action head replaces autoregressive decoding with one- or few-step generation}
10 & GoalFlow \opencode{https://github.com/YvanYin/GoalFlow}~\cite{goalflow2025}              & E2E + goal-cond. flow            & \eLat                  & \bV     \\
11 & WAM-Diff \opencode{https://github.com/fudan-generative-vision/WAM-Diff}~\cite{wamdiff2025}               & Llada-V + MoE + GSPO RL        & \eParam                  & \bV     \\
12 & ViLaD~\cite{vilad2025}                    & Vision-language diffusion        & \eLat                  & \bN     \\
13 & ReCogDrive \opencode{https://github.com/xiaomi-research/recogdrive}~\cite{recogdrive2026}          & Cognitive VLM teacher            & \eLat\,\eParam         & \bV~\bB     \\
14 & ReflectDrive \opencode{https://github.com/pixeli99/ReflectDrive}~\cite{reflectdrive2025}      & Discrete diffusion VLA           & \eLat                  & \bV     \\

\secthrow{C}{Unified Language--Action Codebooks --- shared discrete vocabulary eliminates separate language decoder}
15 & OpenDriveVLA \opencode{https://github.com/DriveVLA/OpenDriveVLA}~\cite{opendrivevla2026} & VLM + discrete actions, 0.5--7B  & \eLat         & \bN     \\

\bottomrule
\end{tabularx}

\vspace{2pt}
{\tiny
\textit{L2 distinguishing property:} language structure is \emph{preserved} at inference
but in a compressed, non-textual form, enabling semantic reasoning without token-by-token LLM decode cost.
}
\end{table*}

\subsection{Conditionally Invoked Language (L3)}
\label{sec:l3}

\begin{table*}[t]
\centering
\scriptsize
\setlength{\tabcolsep}{4pt}
\renewcommand{\arraystretch}{1.3}
\arrayrulecolor{rulecol}

\caption{%
  \textbf{L3 --- Conditionally Invoked Language Methods.}
  Language module is present but executed only when the driving situation warrants it.
  \textbullet\ \faGlobe\ links to an official open-source repository or project page.
  \textbullet\ \textbf{Benchmarks:}
  \bV~NAVSIM~\cciteone{dauner2024navsim},
  \bB~Bench2Drive~\cciteone{jia2024bench2drive},
  \bN~nuScenes~\cciteone{caesar2020nuscenes},
  \bW~Waymo~\cciteone{ettinger2021waymo},
  \bC~CARLA~\cciteone{dosovitskiy2017carla},
  \bP~nuPlan~\cciteone{caesar2021nuplan}.
}
\label{tab:l3_styled}

\rowcolors{2}{white}{stripebg}
\begin{tabularx}{\textwidth}{@{} >{\centering\arraybackslash}p{0.4cm} Y Y Y >{\raggedright\arraybackslash}p{1.8cm} @{}}
\toprule
\textbf{\#} & \textbf{Method} & \textbf{Backbone / Params} & \textbf{Efficiency} & \textbf{Bench.} \\
\midrule

\secthrow{A}{Dual-System Routing --- explicit fast (silent) + slow (language) paths; VLM invoked per frame subset}
1 & LeapAD \opencode{https://github.com/PJLab-ADG/LeapAD}~\cite{leapad2024}             & VLM (Qwen-VL-7B) dual-process + memory       & \eLat            & \bC     \\
2 & DualDriveVLA~\cite{dualdrivevla2026}           & ViT + VLM + scorer              & \eLat   & \bV     \\

\secthrow{B}{Uncertainty-Triggered Invocation --- language module activates when fast path signals insufficient confidence}
3  & FASIONAD~\cite{fasionad2024}                  & ViT + VLM + adaptive feedback   & \eLat            & \bN     \\
4  & FASIONAD++~\cite{fasionadpp2025}              & VAD/GenAD + VLM + uncertainty gate & \eLat            & \bN~\bB \\

\secthrow{C}{Asynchronous \& Adaptive Reasoning --- language tokens generated at variable length or out-of-phase with control}
5 & DriveVLM-Dual~\cite{drivevlm2024}              & Qwen-VL 9.6B + VAD planner        & \eLat            & \bN \\
6 & SOLVE / SOLVE-VLM~\cite{solvevlm2025}          & VLM + E2E network synergy       & \eLat            & \bN     \\
7 & ETA \opencode{https://github.com/opendrivelab/ETA}~\cite{eta2025}                            & LLaVA-Next $\sim$7B + fast ctrl & \eLat   & \bB     \\
8 & AutoVLA \opencode{https://github.com/ucla-mobility/AutoVLA}~\cite{autovla2025}                    & Qwen2.5-VL-3B + GRPO RFT        & \eToken\,\eLat   & \bW~\bP~\bN~\bC \\
9 & AdaThinkDrive~\cite{adathinkdrive2025}        & VLM + adaptive thinking RL      & \eToken          & \bV     \\
10 & ReAL-AD~\cite{realad2025}                     & VLM + human-like multi-stage    & \eLat            & \bN~\bB \\

\bottomrule
\end{tabularx}

\vspace{2pt}
{\tiny
\textit{L3 distinguishing property:} the language module is a run-time component,
but most frames never invoke it. Language cost scales with workload complexity,
not with wall-clock time.
}
\end{table*}

L3 methods preserve language at inference but invoke it selectively: a lightweight
visuomotor module handles routine frames while a heavier language-capable component
activates on scene complexity or model uncertainty.
The central design freedom at L3 is the \emph{trigger}: the mechanism that decides,
per frame, whether language should run at all (Table~\ref{tab:l3_styled}).

\emph{Dual-system routing} (L3-A) uses a learned gate to classify
frames as routine or complex, offering predictable routing overhead but requiring a
reliable complexity proxy~\citep{dualdrivevla2026}.
\emph{Uncertainty-triggered invocation} (L3-B) instead reads the fast
path's confidence score, making the gate data-driven while inheriting that path's
calibration sensitivity~\citep{fasionad2024}.
\emph{Asynchronous and adaptive reasoning} (L3-C) decouples
language generation from the control loop entirely, accepting potential staleness in
exchange for non-blocking actuation~\citep{adathinkdrive2025}.

Among trigger designs, the dual-system router currently offers the best-characterized
performance boundary:
\emph{DualDriveVLA}~\citep{dualdrivevla2026} instantiates the canonical dual-system
structure: a vision-only ViT paired with a VLM invoked on only ${\sim}$15\% of
scenarios via a learned scorer, reaching 91.00~PDMS on NAVSIM~v1 with
a 1.9$\times$ latency speedup over an always-on VLM baseline.
The same paper makes the core trade-off explicit: its \emph{HybridDriveVLA} variant
(classified L4 in our taxonomy, since it runs both pathways on every frame with no
conditional gating) reaches 92.10~PDMS by running both pathways and selecting the
better output, recovering accuracy but forfeiting the efficiency gain, fixing the
frontier that L3 must navigate.

\subsection{Full Language at Inference (L4)}
\label{sec:l4}

\begin{table*}[t]
\centering
\scriptsize
\setlength{\tabcolsep}{4pt}
\renewcommand{\arraystretch}{1.3}
\arrayrulecolor{rulecol}

\caption{%
  \textbf{L4 --- Full Language at Inference.} 
  Language generated every frame; efficiency via smaller models or sparser visual inputs.
  \textbullet\ \textbf{Benchmarks:}
  \bV~NAVSIM~\cciteone{dauner2024navsim},
  \bB~Bench2Drive~\cciteone{jia2024bench2drive},
  \bN~nuScenes~\cciteone{caesar2020nuscenes},
  \bD~DriveLM~\cciteone{drivelm2023},
  \bC~CARLA~\cciteone{dosovitskiy2017carla},
  \bW~Waymo~\cciteone{ettinger2021waymo},
  \bA~PhysicalAI-AV~\cciteone{physicalaiav2025}.
  \textbullet\ \faGlobe\ links to an official open-source repository or project page.
}
\label{tab:l4a_styled}

\rowcolors{2}{white}{stripebg}
\begin{tabularx}{\textwidth}{@{} >{\centering\arraybackslash}p{0.4cm} Y Y Y >{\raggedright\arraybackslash}p{1.8cm} @{}}
\toprule
\textbf{\#} & \textbf{Method} & \textbf{Backbone / Params} & \textbf{Efficiency} & \textbf{Bench.} \\
\midrule

\secthrow{A}{Compact-Architecture VLAs --- small or lightweight backbone reduces parameter / memory cost}
1  & LMDrive \opencode{https://github.com/opendilab/LMDrive}~\cite{lmdrive2024}                   & LLaMA-7B + custom vision encoder                  & \eParam              & \bC     \\
2  & EM-VLM4AD \opencode{https://github.com/akshaygopalkr/EM-VLM4AD}~\cite{emvlm4ad2024}                & Lightweight multimodal VLM      & \eMem\,\eFlops       & \bD     \\
3  & FastDrive~\cite{fastdrive2025}               & 0.9B VLM, structured output     & \eParam\,\eLat       & \bN     \\
4  & DSDrive~\cite{dsdrive2025}                   & Compact LLM + dual-head         & \eParam\,\eLat       & \bC     \\
5  & Drive-KD~\cite{drivekd2026}                  & VLM (InternVL3/Qwen2.5-VL) $\to$ smaller VLM & \eLat\,\eParam   & \bN     \\
6  & MindDrive~\cite{minddrive2025}               & Qwen2-0.5B + dual-LoRA (Decision/Action experts) & \eParam\,\eLat & \bB     \\

\secthrow{B}{Visual-Token Pruning --- reduce LLM input length by selecting informative image tokens}
7  & FastDriveVLA~\cite{fastdrivevla2025}         & Plug-in ReconPruner (MAE)       & \eFlops\,\eLat       & \bN     \\
8  & Prune2Drive \opencode{https://github.com/MinhaoXiong/Prune2Drive}~\cite{prune2drive2025}           & Multi-view diversity pruner     & \eFlops\,\eMem       & \bD     \\
9  & ETA-VLA~\cite{wang2026etavla}                & Text-guided token scoring       & \eFlops       & \bV     \\
10 & ST-Prune~\cite{stprune2026}                  & Spatio-temporal pruner          & \eToken\,\eLat   & \bD     \\

\secthrow{C}{Dynamic Depth \& Early Exit --- adapt computation depth per sample; exit before final layer when confident}
11 & DeeAD~\cite{deead2025}                    & ORION + action-guided exit       & \eLat\,\eFlops      & \bB     \\
12 & Nav-EE~\cite{hu2025navee}                 & VLM + navigation-guided exit     & \eLat               & \bW     \\

\secthrow{D}{Algorithm--System Co-Design --- joint optimization of model and serving stack; quantization and MoE routing included}
13 & DriveMoE~\cite{drivemoe2025}              & MoE VLA (Vision router + 1-shared/6-expert Action MoE)
  & \eToken           & \bB~\bN \\
14 & FlashDriveVLA \opencode{https://github.com/z-lab/flashdrive}~\cite{flashdrivevla2026}    & Alpamayo 1/1.5-10B, full pipeline     &  \eLat\,\eMem     & \bA     \\
15 & LinkVLA~\cite{linkvla2026}                    & InternVL2-1B (InternViT-300M + Qwen2-0.5B) + discrete codebook  & \eLat   & \bC     \\

\bottomrule
\end{tabularx}

\vspace{2pt}
{\tiny
\textit{L4 distinguishing property:} full per-frame language generation is preserved;
efficiency gains come from \emph{within} the language module rather than from reducing when language is used.
}
\end{table*}
L4 methods run the full language pathway on every frame at inference: the language
model participates in perception, reasoning, and action generation with no gate and
no fallback.
Because language runs unconditionally here, the central design freedom at L4 shifts
from \emph{whether} to invoke it to \emph{where to reduce its cost}: the compression
target within an always-on pathway (Table~\ref{tab:l4a_styled}).

\emph{Compact-architecture VLAs} (L4-A) lower fixed inference cost by choosing smaller
backbones from the outset, capping representational capacity in
exchange~\citep{minddrive2025,fastdrive2025}.
\emph{Visual-token pruning} (L4-B) discards uninformative image patches before the LLM
stage, cutting attention complexity but bounded by the reliability of the saliency
estimate~\citep{prune2drive2025}.
\emph{Dynamic depth and early exit} (L4-C) adapt computation per frame, letting routine
inputs leave the network before the final layer while inheriting the calibration
sensitivity of the exit criterion~\citep{deead2025}.
\emph{Algorithm--system co-design} (L4-D) co-optimizes across quantization, prefill
scheduling, and the hardware--software interface, yielding gains that are hardware-bound
rather than portable~\citep{flashdrivevla2026}.

Among these strategies, compact-architecture results speak most directly to the central
L4 question of whether language must be large:
\emph{MindDrive}~\citep{minddrive2025} matches a 7B-LLM VLA on Bench2Drive using a
0.5B backbone, evidence that at L4 the language model can be scaled down without a
proportional accuracy loss. However, evaluation heterogeneity is also highlighted at L4: results rarely share a checkpoint or
protocol, leaving it unclear whether efficiency gains on one axis (e.g., visual-token
pruning) compound with those on another (e.g., early exit or quantization).

\section{NLP/LLM and AD: Migration and Divergence}
\label{sec:nlp-ad}

Core efficiency methods such as quantization, MoE, speculative decoding, PEFT, and distillation stem from LLM prioritizing low token costs and soft latency. Figure~\ref{fig:migration} illustrates how these methods migrate to AD-VLA architectures. Target-agnostic mechanisms transfer seamlessly, whereas methods dependent on conditions specific to LLMs must be adapted to meet real-time autonomous driving constraints.

\noindent\textbf{Migration landscape.}
Distillation is repurposed as reasoning supervision without decoding: a
language-capable teacher shapes a visuomotor student that incurs zero language cost at
runtime~\citep{orionlite2026,dima2025}.
MoE routing is driven by scene context rather than token identity, exploiting
the long-tail distribution of driving situations~\citep{drivemoe2025}.
Speculative decoding inspires fast--slow pipelines where the ``draft'' is a
short-horizon plan or meta-action verified by a slower VLM rather than a token prefix,
giving a fixed dual-frequency architecture in place of the LLM's elastic draft
length~\citep{drivevlm2024}.
Early exit transfers, but its trigger flips from token-confidence to physical
feasibility: a trajectory within prior bounds is the new high-probability token,
and the exit criterion is grounded in safety margin rather than output
entropy~\citep{deead2025}.
Quantization and PEFT/LoRA transfer most directly~\citep{gptq2022,awq2023,lora2021};
visual-token and KV-cache pruning shift the efficiency bottleneck from decode (LLM)
to prefill (AD-VLA)~\citep{stprune2026,prune2drive2025}.
Serving-layer optimizations (e.g., continuous batching, PagedAttention~\citep{vllm2023}) do not transfer: they presuppose batchable
independent requests, structurally absent from AD's single-stream deployment.

\noindent\textbf{Most aggressive reshapings at L1 and L2.}
L1 and L2 represent AD pushing past its LLM teachers.
Distillation collapses into a regime where language is fully erased from the inference
graph rather than merely compressed~\citep{vlp2024,alnp32025,orionlite2026}:
the student inherits no chain-of-thought at runtime, not just a smaller one.
Latent-reasoning work treats register tokens and meta-action symbols as first-class
outputs~\citep{minddrive2025,colavla2025,drivor2026,senna2024} in a way mainstream
LLM systems have only just begun to explore.
The unifying thread: computation gates are \emph{task-structural and externally
grounded}, not model-internal.

Four patterns are either AD-specific or substantially reshaped by its constraints:
\emph{(i)} asynchronous fast--slow inference clocked at distinct fixed
frequencies~\citep{drivevlm2024,eta2025};
\emph{(ii)} meta-action shortcuts collapsing a planning segment into one discrete
token~\citep{senna2024};
\emph{(iii)} train-time-only language supervision fully removed from the inference
graph~\citep{vlp2024,alnp32025};
\emph{(iv)} scene-gated computation conditioned on external driving context rather
than model confidence~\citep{drivemoe2025}.

\noindent\textbf{Reverse influence: a task-structural view of when language helps.}
Selective language invocation has been studied extensively in LLM systems:
model cascades~\citep{cascade2023}, adaptive and budget-controlled
CoT~\citep{adaptivecot2025}, confidence-based early exit~\citep{calm2022},
speculative decoding~\citep{leviathan2023specdecode,eagle}, sparse MoE
routing~\citep{shazeer2017moe,mixtral2024}, and retrieval/tool gating in agent
systems.

What distinguishes AD is not \emph{whether} language is gated, but \emph{what the
gate conditions on}.
LLM-system gating is predominantly driven by model-internal signals (e.g. token
confidence, output entropy, learned reasoning budgets) under soft latency targets.
AD's hard-deadline, single-stream setting forces gates conditioned on
\emph{external task structure}: scene complexity, agreement with a planning prior,
dynamics-conditioned thresholds, and asynchronous clocks tied to physical control
rates.
We suggest this task-structural view of conditional language computation is the
transferable insight in the AD$\rightarrow$LLM-systems direction, particularly for
latency-critical applications
where gating signals can be grounded in the environment or task contract rather than
in the language model's own uncertainty.
The pattern, taken together, is that AD adopts LLM efficiency techniques where they
accelerate computation but invents its own where the mechanism itself imposes language load, and the Language Residue taxonomy is what makes the asymmetry legible.

\section{Benchmarks and Evaluation}
\label{sec:benchmarks}
\begin{figure}
    \centering
    \includegraphics[width=1\linewidth]{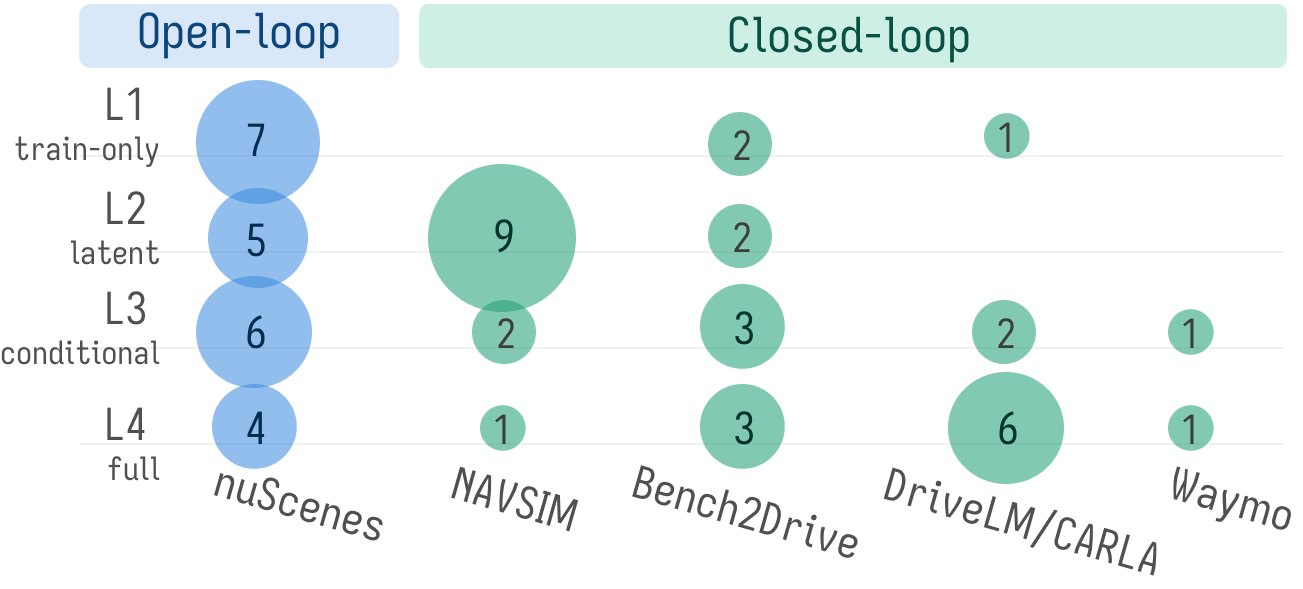}
    \caption{%
  {Main Benchmark Coverage by Language Residue Taxonomy.} Each cell reports the number of methods at that level with published results
  on the named benchmark. A method may appear in multiple columns. Only widely used benchmarks are plotted.}
    \label{fig:bench-coverage}
    \vspace{-1em}
\end{figure}

\subsection{Open-Loop Benchmarks}

\noindent\textbf{nuScenes}~\citep{caesar2020nuscenes} measures trajectory-prediction
accuracy over 1{,}000 real-world driving scenes (1.4M camera frames), using L2
displacement error (m) and collision rate (\%) as primary metrics.
It is the most-reported benchmark in this survey (Figure~\ref{fig:bench-coverage}), though its use of
ego-status as a model input has drawn sustained criticism for leaking ground-truth
velocity information.

\noindent\textbf{NAVSIM}~\citep{dauner2024navsim} scores open-loop trajectories against
a non-reactive closed-loop oracle, collapsing multiple driving competencies into a single
Predictive Driver Model Score (PDMS; v2: EPDMS).
It is the primary substrate for L2 methods, and supplanted nuScenes as the standard open-loop evaluation platform
from 2025 onward.

\subsection{Closed-Loop Benchmarks}

\noindent\textbf{Bench2Drive}~\citep{jia2024bench2drive} tests interactive closed-loop
driving across 44 routes and 220 CARLA scenarios, measuring Driving Score (DS) and
Success Rate (SR) across multi-ability test conditions.

\noindent\textbf{CARLA Leaderboard~2.0} extends the CARLA simulator~\citep{dosovitskiy2017carla} to long routes (${\sim}$10\,km) with over 20 hazard types, reporting DS,
route completion, and infraction counts.

\noindent\textbf{DriveLM}~\citep{drivelm2023} evaluates language-grounded reasoning
through a Graph Visual Question Answering (GVQA) protocol that chains perception, prediction, and planning over nuScenes and CARLA scenes (${\sim}$440K QA pairs). It reports a composite Final Score that aggregates accuracy on
multiple-choice questions, a GPT-judged score for planning answers,
language-generation metrics (BLEU, ROUGE-L, CIDEr) for perception, and an
object-matching score for prediction.

\noindent\textbf{Waymo Open Motion Dataset (WOMD)}~\citep{ettinger2021waymo} benchmarks
long-horizon trajectory prediction and planning over 103K real-world scenes, reporting
minimum average displacement error (minADE) and minimum final displacement error (minFDE).

\section{Open Problems}
\label{sec:saturation}

\subsection{Trustworthiness and Robustness}

{\ding{73}} \textit{Out-of-distribution (OOD) is underexplored.} Language models are typically kept on the inference path specifically to navigate the long tail \citep{alpamayo2025,corevla2025}. However, as \citet{impromptuvla2025} note, driving VLAs still fail in unstructured corner cases due to a scarcity of targeted benchmarks. This creates a paradox: the component tasked with absorbing distribution shifts sits precisely where alignment data is thinnest. Conditional invocation systems \citep{eta2025,autovla2025,adathinkdrive2025} turn this gap into a functional risk. By dynamically gating the language path, they implicitly guess whether a scene falls outside the fast-path distribution, without validating those choices against any shift criterion. The irony is that the required machinery already exists. Formulations for runtime OOD detection \citep{guo2026ood} and independent behavior benchmarks \citep{distelzweig2026behaviorbench} are both available, yet entirely absent from current evaluations.
\vspace{5pt}
\noindent {\ding{73}} \textit{Hallucination is introduced by design.}
Language supervision rewards consistency with language priors rather than factual observation. Thus, unpenalized misreadings easily poison downstream planning. \citet{fan2024hcoenet} confirm this stems from the objective itself, reporting massive F1 gains when cross-checking is enforced. Crucially, this failure requires no adversarial trigger, making it a common operational risk that security-focused research largely ignores. Mitigation strategies target different points of intervention: reinforcement learning evaluates the reasoning trace by driving outcomes \citep{driver12025,autodrive2025}; auxiliary objectives anchor semantic outputs to geometric predictions \citep{occvla2025,alnp32025}; and runtime supervision filters narrations before they reach the controller \citep{li2026csn}. Because these methods lack a unified hallucination metric, their comparative effectiveness remains unknown.

\vspace{5pt}
\noindent {\ding{73}} \textit{Attacks reach action through reasoning.} Driving VLMs face severe vulnerabilities across both white-box \citep{zhang2026advlm} and black-box \citep{wang2025cad} regimes, where corrupting the reasoning process directly corrupts the trajectory. Conditional invocation introduces a secondary target: because the gate defines a learnable mapping from an observable condition to a behavioral change, adversaries can trigger malicious behavior through common physical objects \citep{ni2024badvlmdriver} or achieve 90\% attack success at a mere 10\% poisoning ratio \citep{wang2026gla}. 
Countermeasures remain partial and lack level-agnostic guarantees. Pre-deployment evaluation tools like scenario generation and fault injection \citep{safebench2022,pasandideh2026llmfault} target the general driving stack rather than the language path specifically. At runtime, supervision over emitted narrations \citep{li2026csn} and distribution-shift detection \citep{guo2026ood} serve to detect and degrade rather than prevent attacks, with the former requiring an explicitly exposed language channel. On the model side, the sole VLA attack-defense framework is limited to manipulation policies \citep{xu2025edpa}, and broader defense mechanisms \citep{li2026vlasafety} are completely unevaluated for driving. 

\subsection{Benchmark and Evaluation}
{\ding{73}} \textit{Efficiency--safety frontier is unquantified.}
Most work reports latency or FLOPs reductions without measuring whether reduced
compute raises infraction rates under edge-case scenarios. This gap is
confirmed by \citet{safebench2022}, who find that autonomous driving algorithms
exhibit a direct performance trade-off between benign and safety-critical
scenarios, a trade-off that is rarely characterized jointly in efficiency-focused
papers. Moreover, \citet{pasandideh2026llmfault} demonstrate that under sensor-degradation
faults such as fog, RMSE can increase by up to 99\% relative to clean-data
baselines, yet such stress conditions are almost never included in
compute-efficiency evaluations.

\vspace{5pt}
\noindent
{\ding{73}} \textit{Benchmark fragmentation prevents cross-level comparison.}
Only a small fraction of surveyed methods report on both a trajectory-planning
benchmark and a language/reasoning benchmark, leaving open whether efficiency
gains in planning come at the cost of reasoning capability. This fragmentation
is structural: trajectory benchmarks (nuPlan, NAVSIM, Bench2Drive) and
language-QA or scene-understanding benchmarks (DriveLM, DriveCombo, VLM safety
suites) are evaluated by disjoint communities with incompatible metrics.
\citet{distelzweig2026behaviorbench} argue that the most widely-used planning benchmark,
nuPlan Val14, ``can be largely solved by lane-following combined with simple
collision-checking,'' meaning high scores do not imply genuine reasoning
capability. Conversely, \citet{ma2026drivecombo} show that state-of-the-art
multimodal models degrade sharply on traffic-rule conflict tasks (L5 of their
five-level cognitive ladder) despite strong trajectory-level numbers. No
existing work bridges these two evaluation axes in a single controlled study.

\subsection{Deployment and Future Gap}
{\ding{73}} \textit{The L3 trigger design problem has no principled solution.}
Adaptive-compute systems must decide \emph{when} to engage deliberative
reasoning rather than reactive control, yet no criterion distinguishes
deterministic clocks, entropy thresholds, and learned gates, and the safety
implications of trigger choice remain unexplored. AdaThinkDrive
\citep{adathinkdrive2025} recently demonstrated that a VLA framework with a
``Fast answering / Slow thinking'' mechanism can reduce inference time by 14\%
on NAVSIM, but the trigger design is evaluated purely for efficiency; the safety
consequences of misclassifying a scene as ``easy'' are not examined.

\vspace{5pt}
\noindent
{\ding{73}} \textit{L2 latent reasoning resists inspection.}
Efficiency gains from compressing cognition into latent representations come at
the cost of interpretability. A comprehensive survey of world models for
autonomous driving \citep{feng2025worldmodelsurvey} identifies ``opaque latent
policies'' as one of three persistent open issues alongside generalization and
formal safety guarantees.

\vspace{5pt}
\noindent
{\ding{73}} \textit{Real-world deployment evidence is sparse.}
The large majority of surveyed methods are validated exclusively in simulation,
leaving the sim-to-real transfer of both efficiency and safety properties
unknown. \citet{hu2024sim2realsurvey} identify differences in lighting, textures,
vehicle dynamics, and agent behavior as persistent sources of the reality gap.
\citet{arango2026sim2realplatform} find that discrepancies in ego-vehicle dynamics,
sensor behavior, and processing times introduce a performance gap that
survives domain-randomization and digital-twin transfer.

\section{Conclusion}
\label{sec:conclusion}

We organized the emerging design space of efficient VLA for autonomous
driving around a single question: how much language computation should survive
to inference?
The Language Residue taxonomy (L1--L4) provides a principled answer space,
classifying representative methods by their inference-time language
involvement and annotating each across five deployment axes (Latency, Parameters, Memory, FLOPs, Tokens).
Analysis of the NLP/LLM--AD efficiency genealogy reveals a selective migration: techniques
that accelerate computation transfer with adaptation, while techniques whose mechanism
is language-load are replaced by AD-native designs grounded in task structure.
Cross-level benchmark analysis exposes fragmentation as a primary barrier to fair
comparison and efficient progress.
We hope the taxonomy and the continuously updated repository at \href{https://github.com/iamtongfei/Awesome-Efficient-VLA4AD}{\texttt{GitHub}}
serve as shared vocabulary for the community and accelerate progress toward efficient,
safe, and interpretable autonomous driving.

\section*{Limitations}
Several limitations are worth noting.
Coverage is weighted toward methods in which a language module sits on the
inference path, since the Language Residue levels are defined by that module's
presence. Earlier language-conditioned and modular driving systems are cited for
context but not assigned levels, as they do not expose a comparable inference-time
language budget.
Concurrent work not yet publicly available is also outside our coverage.
Benchmark analysis is constrained by what authors self-report: efficiency metrics
(FLOPs, latency, memory) and driving performance metrics rarely appear in the same
paper, limiting our ability to characterize the efficiency--accuracy Pareto frontier.
The large majority of surveyed methods are evaluated in simulation; real-world
deployment results are available for only a handful of systems~\citep{drivevlm2024},
leaving the simulation-to-real transfer of efficiency claims uncharacterized.
Reported gains are platform-bound: latency and memory hold only for the
hardware measured, while parameter counts, FLOPs, and token counts transfer
across platforms. Few methods report timings on automotive-grade
hardware, so speedups should be read as relative.
The taxonomy assigns each method a single primary Language Residue level, which may
oversimplify architectures that combine, for example, L1-style pretraining with
L3-style conditional dispatch at inference.

\bibliography{references}
\appendix

\end{document}